\documentclass{article}
\usepackage{iclr2027_conference,times}

\usepackage{amsmath,amsfonts,bm}

\def\eqref#1{equation~\ref{#1}}

\def\1{\bm{1}}

\DeclareMathAlphabet{\mathsfit}{\encodingdefault}{\sfdefault}{m}{sl}
\SetMathAlphabet{\mathsfit}{bold}{\encodingdefault}{\sfdefault}{bx}{n}

\def\sR{{\mathbb{R}}}

\usepackage{hyperref}
\usepackage{url}
\usepackage{amsmath,amssymb}
\usepackage{graphicx}
\usepackage{booktabs}
\usepackage{multirow}
\usepackage{subcaption}
\usepackage{float}
\usepackage{caption}

\newcommand{\bx}{\mathbf{x}}

\newcommand{\by}{\mathbf{y}}

\iclrfinalcopy

\title{Recovering Governing Dynamics from Distributed Observations via Exact Spline Merging}

\author{Naveen Mysore \\
University of California, Santa Barbara \\
Dyssonance AI \\
\texttt{nmysore@ucsb.edu} \\ \texttt{nmysore.work@gmail.com}}

\begin{document}
\maketitle
\thispagestyle{fancy}
\lhead{Under review as a conference paper at ICLR 2027}
\begin{abstract}
Scientific observations are frequently distributed across locations, time periods, and institutions. Combining such observations into a continuous, differentiable field enables recovering governing physical parameters from its derivatives. This paper makes two contributions in this setting. First, the established additive structure of fixed-basis ridge-regression statistics is applied to tensor-product spline fields: each data holder computes a local Gram matrix and moment vector, and the merged solution is mathematically identical to centralized fitting, with no raw data shared and no iterative synchronization. This property is specific to the fixed-feature squared-error setting; the present derivation does not establish an analogous guarantee for general jointly trained multilayer networks. Second, a complete pipeline connects distributed observations to physical parameter inference through field reconstruction, derivative extraction, and linear regression. The pipeline is validated on four PDEs: diffusion, wave, heat-with-source, and the nonlinear viscous Burgers equation, recovering governing parameters to sub-percent accuracy in the linear cases and 5\% for Burgers. In all cases, distributed merging introduces zero degradation relative to centralized fitting. Synthetic experiments validate parameter recovery; application to 41 years of NOAA sea-surface temperature data validates field reconstruction and aggregation equivalence on real spatiotemporal observations. Source code to reproduce all experiments is available at \url{https://github.com/NAVEENMN/splinemerge}.
\end{abstract}

\section{Introduction}

\begin{figure}[t]
\begin{center}
\includegraphics[width=\linewidth]{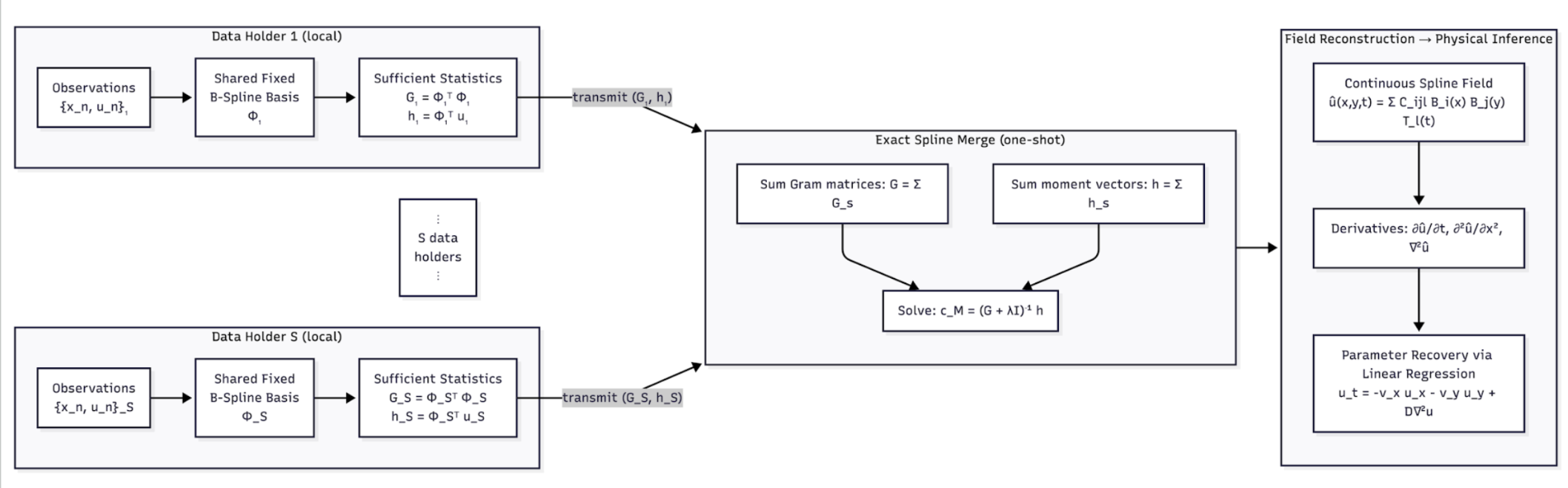}

\vspace{0.3em}

\begin{minipage}[t]{0.49\linewidth}
\centering
\includegraphics[width=\linewidth]{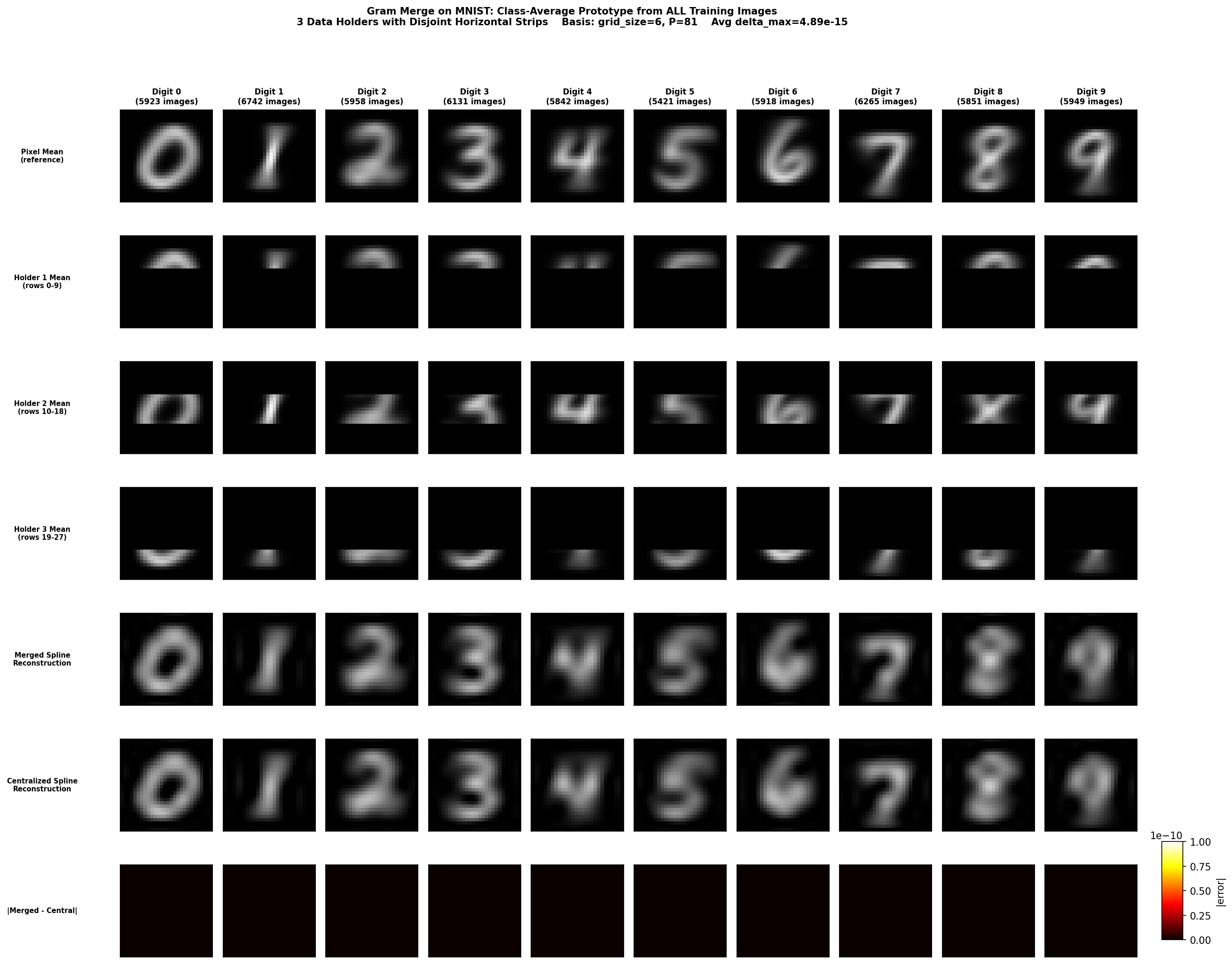}
\end{minipage}
\hfill
\begin{minipage}[t]{0.49\linewidth}
\centering
\includegraphics[width=\linewidth]{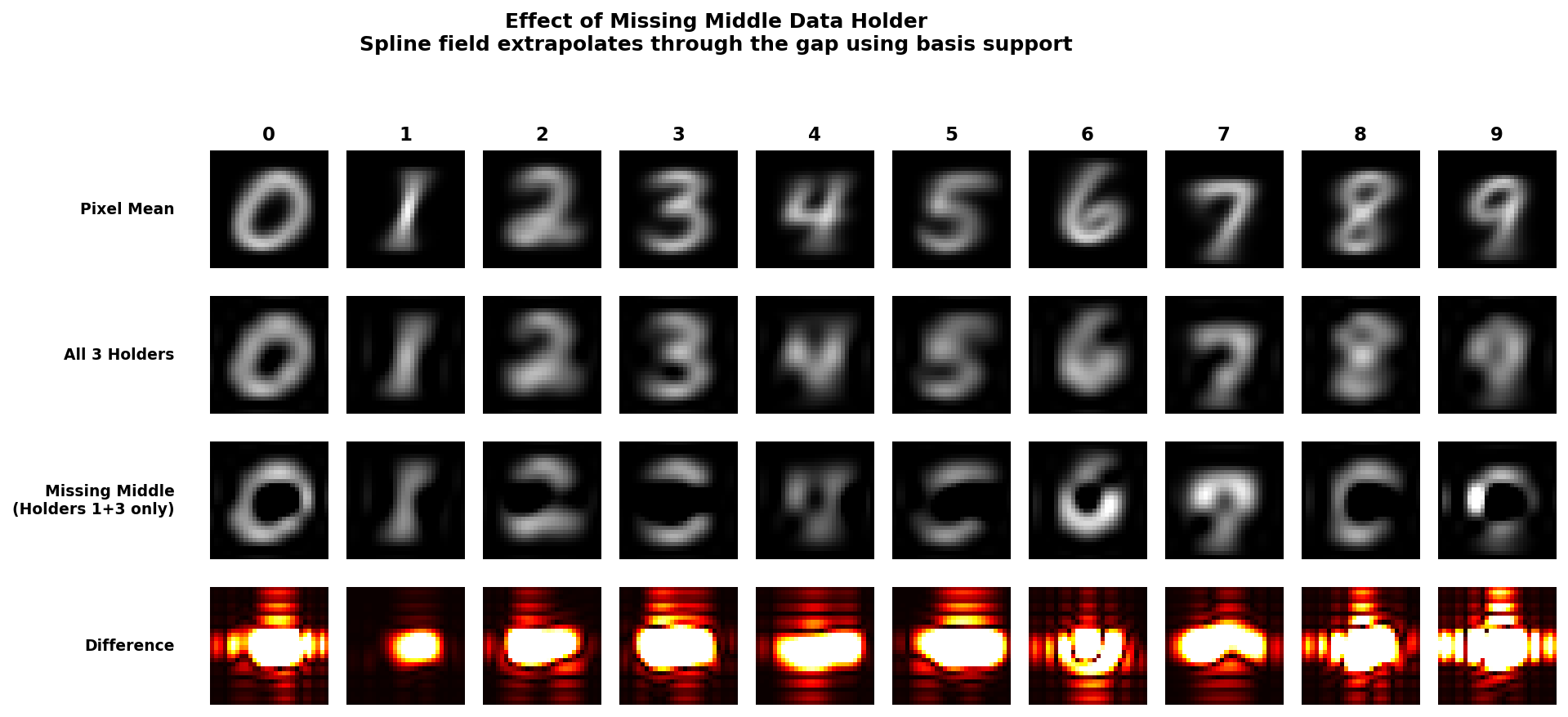}
\end{minipage}
\end{center}
\caption{\textbf{Top:} End-to-end pipeline. Each data holder computes sufficient statistics $(G_s, h_s)$ from local observations using a shared fixed B-spline basis and transmits only these (not the raw data). A single matrix solve produces the exact centralized solution; the resulting continuous field yields derivatives for parameter estimation. \textbf{Bottom left:} Demonstration on MNIST. Each 28$\times$28 image is a 2D field $u(x,y)$ fitted with tensor-product B-splines ($P{=}81$). Three data holders each observe a disjoint horizontal strip (rows 0--9, 10--18, 19--27) across all ${\sim}6{,}000$ training images per digit class. The merged spline reconstruction is identical to the centralized fit (worst-case $\delta_{\max} < 10^{-13}$ across all ten digits), reconstructing class-average prototypes from ${\sim}1.5$M pixels per holder per digit class. \textbf{Bottom right:} When the middle holder is absent, the spline extrapolates through the gap. Digits with distinctive middle features (3, 4, 6, 8) degrade; simpler shapes (0, 1) remain recognizable. This demonstrates exact merging when all holders contribute and graceful degradation when observations are incomplete.}
\label{fig:overview}
\end{figure}

Scientific observations are often distributed across space, time, and institutions. No single observer has access to the full field, yet recovering governing dynamics benefits from a continuous, differentiable field representation. For models linear in their parameters, specifically fixed-basis spline expansions, the distributed merge problem has an exact, one-shot solution: each data holder computes a local Gram matrix $G_s = \Phi_s^\top \Phi_s$ and moment vector $h_s = \Phi_s^\top \by_s$, and the merged solution $c^* = (\sum_s G_s + \lambda I)^{-1} \sum_s h_s$ is mathematically identical to centralized fitting. No raw data is shared. No iterative synchronization is needed. Figure~\ref{fig:overview} illustrates the pipeline and demonstrates it on MNIST images.

This additive structure of fixed-feature ridge-regression statistics is classical \citep{hoerl1970ridge} and has been explicitly formulated as one-shot federated aggregation by \citet{alsulaimawi2026oneshot}. Related model merging methods \citep{mcmahan2017fedavg,wortsman2022modelsoups,ainsworth2023gitrebasin,jin2023regmean} and KAN merging \citep{polar2025kanmerge} of Kolmogorov--Arnold Networks \citep{liu2024kan} address different settings. Federated scientific machine learning \citep{zhang2025fedsciml,shukla2021parallelpinn} requires iterative training. Field reconstruction methods \citep{raissi2019pinn,santos2023senseiver} and equation discovery \citep{brunton2016sindy,rudy2017pdefind,messenger2021weaksindy} typically assume centralized data access.

Two distinct estimation problems are involved. The first recovers field coefficients from distributed observations; the exact aggregation guarantee applies here. The second estimates physical parameters (diffusivity, wave speed, viscosity) from derivatives of the reconstructed field; this step is approximate. Section~\ref{sec:background} develops both constructions from first principles.

This paper makes two contributions: \textbf{(1) Distributed field reconstruction}: the merge reproduces the centralized estimator on synthetic PDEs, MNIST images, and real climate data. \textbf{(2) End-to-end physics recovery}: a complete pipeline recovers governing parameters across linear and nonlinear PDEs to sub-percent accuracy (5\% for nonlinear Burgers), with a systematic noise sensitivity analysis.

\section{Background}
\label{sec:background}

This section builds the core framework from first principles: basis function representations, the least-squares fitting problem, sufficient statistics, and why they compose exactly under distributed merging.

\textbf{Problem setup.} Multiple data holders observe the same scalar field in a common coordinate system. Each holder retains its own observation locations and measured values. Before fitting, all holders agree on the same fixed basis functions, coefficient ordering, and regularized fitting objective. The first goal is to recover the coefficients that would be obtained by pooling all observations and fitting one model. The second goal is to use the reconstructed field to estimate parameters of a specified governing equation. The term ``data holder'' is used throughout; in the experiments, a data holder may represent a spatial region, an institutional partition, or a class-conditional subset.

\subsection{B-Spline Basis Functions}

A B-spline basis function $B_k(x)$ is a piecewise polynomial defined over a sequence of knot points \citep{deboor2001splines}. For cubic B-splines (degree 3) on a uniform grid with simple interior knots, each basis function has support over four adjacent knot intervals and is zero outside this region. This property, called \emph{compact support}, means that each $B_k$ responds only to nearby data. The resulting spline is $C^2$-continuous (twice continuously differentiable) across interior knots, and the complete basis forms a partition of unity on its base interval.

A scalar field can then be represented as a weighted sum of these basis functions:
\begin{equation}\label{eq:1d_field}
u(x) = \sum_{k=1}^{K} c_k\, B_k(x).
\end{equation}
The representation can express nonlinear dependence on $x$, while remaining linear in its coefficients $c = [c_1, \ldots, c_K]^\top$. This distinction is the foundation of everything that follows.

\subsection{Fitting as Linear Regression}

Given $N$ observations $\{(x_n, y_n)\}_{n=1}^N$, where $y_n$ denotes the measured value at location $x_n$, the goal is to find coefficients $c$ such that $u(x_n) \approx y_n$. Define the \emph{feature matrix} $\Phi \in \sR^{N \times K}$ with entries $\Phi_{nk} = B_k(x_n)$, so that the model predictions are $\hat{\by} = \Phi\, c$. The optimal coefficients minimize the regularized squared error:
\begin{equation}\label{eq:loss}
\mathcal{L}(c) = \|\, \by - \Phi\, c\,\|^2 + \lambda \|c\|^2,
\end{equation}
where $\lambda > 0$ is a ridge regularization parameter \citep{hoerl1970ridge}. With $\lambda > 0$, the objective is strictly convex (the Hessian $2(\Phi^\top\Phi + \lambda I)$ is positive definite even when $\Phi$ is rank-deficient), ensuring a unique global minimum found by setting the gradient to zero:
\begin{equation}
\frac{\partial \mathcal{L}}{\partial c} = -2\,\Phi^\top(\by - \Phi\, c) + 2\lambda\, c = 0.
\end{equation}

Rearranging yields the \emph{normal equations}:
\begin{equation}\label{eq:normal_simple}
(\Phi^\top \Phi + \lambda I)\, c^* = \Phi^\top \by.
\end{equation}

\subsection{Gram Matrix, Moment Vector, and Sufficient Statistics}

The solution in Eq.~(\ref{eq:normal_simple}) depends on the observations only through two quantities:
\begin{equation}\label{eq:sufficient}
G = \Phi^\top \Phi \in \sR^{K \times K}, \qquad h = \Phi^\top \by \in \sR^K.
\end{equation}
The \emph{Gram matrix} $G$ captures how the basis functions co-activate across the observation points: entry $G_{ij} = \sum_n B_i(x_n)\, B_j(x_n)$. The \emph{moment vector} $h$ captures how each basis function correlates with the observed values: $h_k = \sum_n B_k(x_n)\, y_n$.

For the fixed-basis ridge-regression objective in Eq.~(\ref{eq:loss}), the pair $(G, h)$ is sufficient to compute $c^*$: the coefficient estimate depends on the observations only through these two summaries. Once computed, the raw observations $\{(x_n, y_n)\}$ can be discarded without affecting the estimate. This sufficiency is specific to the chosen fitting objective; changing the basis, the loss function, or adding new analyses (such as computing the residual sum of squares, which requires $\by^\top \by$) may require information not preserved in $(G, h)$. For a fixed basis, $N$ observations reduce to a $K \times K$ matrix and a $K$-vector, with the summary size independent of observation count. The term ``sufficient statistics'' is used here in the computational sense of sufficient for the specified ridge estimator; under a Gaussian linear observation model with known noise variance, the same quantities also arise as classical statistical sufficient statistics (joint inference with unknown variance additionally requires $\by^\top\by$).

\subsection{Composability: From Sufficiency to Exact Distributed Merging}

Sufficiency alone is useful for compression. Composability makes it useful for distributed computing.

Suppose observations come from two data holders, $A$ and $B$, at different locations. If all data were pooled, the stacked feature matrix and observation vector would give normal equations:
\begin{equation}
\left(\begin{bmatrix}\Phi_A \\ \Phi_B\end{bmatrix}^\top \begin{bmatrix}\Phi_A \\ \Phi_B\end{bmatrix} + \lambda I\right) c^* = \begin{bmatrix}\Phi_A \\ \Phi_B\end{bmatrix}^\top \begin{bmatrix}\by_A \\ \by_B\end{bmatrix}.
\end{equation}
Expanding the block products:
\begin{equation}\label{eq:merge}
(\underbrace{\Phi_A^\top \Phi_A}_{G_A} + \underbrace{\Phi_B^\top \Phi_B}_{G_B} + \lambda I)\, c^* = \underbrace{\Phi_A^\top \by_A}_{h_A} + \underbrace{\Phi_B^\top \by_B}_{h_B}.
\end{equation}
The sufficient statistics are \emph{additive}. Each data holder computes its local $(G_s, h_s)$ independently and transmits only these compact summaries. The merged solution
\begin{equation}\label{eq:merge_final}
c^*_M = (G_A + G_B + \lambda I)^{-1}(h_A + h_B)
\end{equation}
is identical to the solution that would be obtained by fitting on all observations centrally. No raw data is shared. No iterative synchronization is needed. The merge requires a single matrix solve.

This extends to any number of data holders $S$ by induction: $c^*_M = (\sum_{s=1}^S G_s + \lambda I)^{-1} \sum_{s=1}^S h_s$.

\textbf{Two merge variants.} The formula above reconstructs the centralized \emph{data} fit, using the moment vectors $h_s = \Phi_s^\top \by_s$ that summarize raw observations. A related variant preserves trained \emph{model predictions} instead: if data holder $s$ has already fitted local coefficients $c_s$, the prediction-matching merge is $c_M = (\sum_s G_s + \lambda I)^{-1} \sum_s G_s c_s$. The two coincide when each local fit is at its unregularized optimum ($G_s c_s = h_s$). The experiments in this paper use the data merge.

The summary size per data holder is $O(K^2)$ for the symmetric Gram matrix plus $O(K)$ for the moment vector, independent of observation count. For an ordered univariate degree-$p$ B-spline basis, compact support yields a Gram matrix with half-bandwidth at most $p$. Tensor-product representations induce different sparsity patterns (Section~\ref{sec:method}). Note that dense summaries can exceed the raw data size when $K$ is large relative to $N$; sparsity exploitation or structured transmission would be needed for a practical bandwidth advantage.

\textbf{Fixed versus learned features.} The aggregation derived above applies whenever the feature map is fixed and the coefficients enter a quadratic objective. This includes spline expansions, Fourier bases, and linear readouts on frozen neural representations. A fully trainable multi-layer network generally does not satisfy this condition: changing an earlier layer changes the features supplied to subsequent layers, so the Gram matrix from one parameter setting is no longer valid at another. The present construction therefore does not extend to exact whole-network merging in that setting.

\textbf{From aggregation to physical inference.} The preceding derivation concerns the first estimation problem: recovering field coefficients from distributed observations. Exact aggregation removes any discrepancy between distributed and centralized fitting. It does not, by itself, remove field-approximation error (the basis may not perfectly represent the true field), derivative-estimation error (finite differences or analytic derivatives of an approximate field), or identifiability issues in the physical regression (ill-conditioned derivative ratios). Section~\ref{sec:method} specifies the field representation and the parameter-recovery procedure; the experiments distinguish aggregation discrepancy from reconstruction and parameter-estimation errors, although a complete decomposition of individual error contributions has not been performed.

\section{Method}
\label{sec:method}

The dimensionality of the input does not change the aggregation algebra. A 1D spline curve $u(x) = \sum_i c_i B_i(x)$, a 2D surface $u(x,y) = \sum_{i,j} C_{ij} B_i(x) B_j(y)$, and a 3D spatiotemporal field all reduce to the same object: $f(\bx) = \phi(\bx)^\top c$, where $\phi$ is a fixed feature vector and $c$ is the coefficient vector to be estimated. The Gram merge applies identically in each case (Figure~\ref{fig:overview}). This section specifies the field representation, basis geometry, and parameter-recovery procedure.

\subsection{Tensor-Product Spatiotemporal Fields}

A scalar spatiotemporal field $u: \sR^d \times \sR \to \sR$ is represented using a tensor-product expansion. For a 2D spatial domain with one temporal dimension:
\begin{equation}\label{eq:3d_field}
u(x, y, t) = \sum_{i=1}^{K_x} \sum_{j=1}^{K_y} \sum_{l=1}^{K_t} C_{ijl}\, B_i(x)\, B_j(y)\, T_l(t),
\end{equation}
where $B_i$ are cubic B-spline basis functions and $T_l$ are temporal basis functions (either B-splines or Fourier harmonics). Defining the tensor-product feature $\phi_{ijl}(x,y,t) = B_i(x)\, B_j(y)\, T_l(t)$ and flattening the coefficients into a vector $c = \mathrm{vec}(C) \in \sR^P$ where $P = K_x \cdot K_y \cdot K_t$, the field becomes $u = \phi^\top c$. For $N$ observations, the feature matrix $\Phi \in \sR^{N \times P}$ has entries $\Phi_{n,(i,j,l)} = B_i(x_n)\, B_j(y_n)\, T_l(t_n)$, and the Gram merge from Section~\ref{sec:background} applies directly.

B-spline basis functions are evaluated using the piecewise polynomial evaluation described by \citet{mysore2026inkan}, implemented in the publicly available InKAN package (v0.4.2). All experiments are implemented in PyTorch \citep{paszke2019pytorch} with NumPy \citep{harris2020numpy} and SciPy \citep{virtanen2020scipy} for numerical routines.

\subsection{Basis Geometry and Merge Statistics}

Basis support constrains the possible sparsity of the empirical feature-interaction statistics. For local B-splines (degree $p$), each $B_i(x)$ has compact support spanning at most $p+1$ knot intervals, and the resulting Gram matrix is banded within each univariate block. For tensor-product representations, an interaction is zero when the basis supports fail to overlap in any coordinate. Global features such as $\cos(m\omega x)$ produce generally non-sparse Gram matrices whose conditioning depends on frequency selection, observation coverage, and weighting.

The actual entries, conditioning, and behavior of approximate merging methods additionally depend on the represented function space, regularization, and local fitting accuracy. Basis support alone is insufficient as a causal explanation for averaging performance. The present experiments demonstrate exact statistic aggregation for spline and hybrid representations; a controlled basis-family comparison is left to future work.

The NOAA experiment (Section~\ref{sec:noaa}) uses a hybrid representation: spatial B-splines crossed with temporal Fourier harmonics and polynomial trend terms, producing a tensor-product model with spatially varying seasonal amplitudes.

\subsection{Physical Parameter Recovery}

Given the fitted continuous field $u(x,y,t)$, spatial and temporal derivatives are computed via central finite differences on a regular evaluation grid (interior points only). Physical parameters that appear linearly in a specified PDE are then recovered by ordinary least squares. Let $\ell_i$ denote the computed Laplacian $\nabla^2 u$ at interior grid point $i$. For the diffusion equation $u_t = D\nabla^2 u$, the recovered diffusion coefficient is:
\begin{equation}
D_{\mathrm{rec}} = \frac{\sum_i \ell_i\, u_{t,i}}{\sum_i \ell_i^2}.
\end{equation}
For the wave equation $u_{tt} = c_w^2 \nabla^2 u$, the analogous formula yields $c_{w,\mathrm{rec}}^2 = \sum_i \ell_i\, u_{tt,i}\, /\, \sum_i \ell_i^2$, with $c_{w,\mathrm{rec}} = \sqrt{c_{w,\mathrm{rec}}^2}$ when the recovered squared value is non-negative and treated as invalid otherwise. For the inhomogeneous heat equation $u_t = D\nabla^2 u + S$, the known source is subtracted: $D_{\mathrm{rec}} = \sum_i \ell_i\,(u_{t,i} - S_i)\, /\, \sum_i \ell_i^2$. For Burgers $u_t + u\,u_x = \kappa\, u_{xx}$, the viscosity is recovered as $\kappa_{\mathrm{rec}} = \sum_i u_{xx,i}\,(u_{t,i} + u_i\, u_{x,i})\, /\, \sum_i u_{xx,i}^2$.

This is the second problem described in Section~\ref{sec:background}: the physical parameters are inferred from the reconstructed field, not from raw observations. The accuracy depends on field-approximation quality, derivative computation, and conditioning of $\sum_i \ell_i^2$ (a numerical stabilizer of $10^{-8}$ is added to the denominator in the implementation). When the Laplacian is nearly zero everywhere, the denominator is small and the recovery is unreliable. Because the computed Laplacian contains finite-difference truncation error, the OLS regression operates with noisy features. The resulting bias direction is not fixed: errors in the response ($u_t$) and features ($\ell$) are derived from the same fitted field and can be correlated, so the bias may have either sign.

\textbf{Remark on local rank deficiency.} Under spatial fragmentation, a data holder's local Gram matrix $G_s$ is typically rank-deficient for compactly supported B-splines (basis functions outside the observed region contribute zero columns). However, all exact unregularized local least-squares minimizers produce the same product $G_s c_s = h_s$, so local uniqueness of $c_s$ is unnecessary for the data merge. The composability guarantee extends to any globally shared quadratic penalty $\lambda\, c^\top R\, c$ (e.g., derivative-based smoothing penalties), provided $\sum_s G_s + \lambda R$ is positive definite.

\section{Experiments}

The pipeline is validated on four synthetic PDEs (three linear, one nonlinear) with known ground truth and on real climate data. All PDE experiments use cubic B-splines ($p{=}3$) with ridge regularization $\lambda = 10^{-4}$ and float64 arithmetic. Observations are sampled uniformly at random; distributed merge experiments split the domain spatially, with equal observations per data holder. Synthetic reference solutions use analytic Fourier-mode evolution (diffusion, wave, heat-with-source) or pseudospectral integration (Burgers). Full experimental configuration is in Appendix Table~\ref{tab:config}.

\subsection{Diffusion Equation}

The diffusion equation $u_t = D(u_{xx} + u_{yy})$ with true diffusivity $D = 0.05$ is the simplest PDE in the pipeline: one governing parameter, first-order in time. Each Fourier mode evolves analytically, decaying as $\hat{u}(\mathbf{k}, t) = \hat{u}(\mathbf{k}, 0)\, e^{-D|\mathbf{k}|^2 t}$. From 10,000 random spatiotemporal observations, a 3D tensor-product B-spline field ($K_x{=}11, K_y{=}11, K_t{=}11$; $P{=}1{,}331$) is fitted and the diffusion coefficient is recovered via single-parameter regression: $D_{\mathrm{rec}} = 0.04994$ (0.12\% relative error from unrounded value).

\begin{figure}[t]
\begin{center}
\begin{subfigure}[b]{\linewidth}
\includegraphics[width=\linewidth]{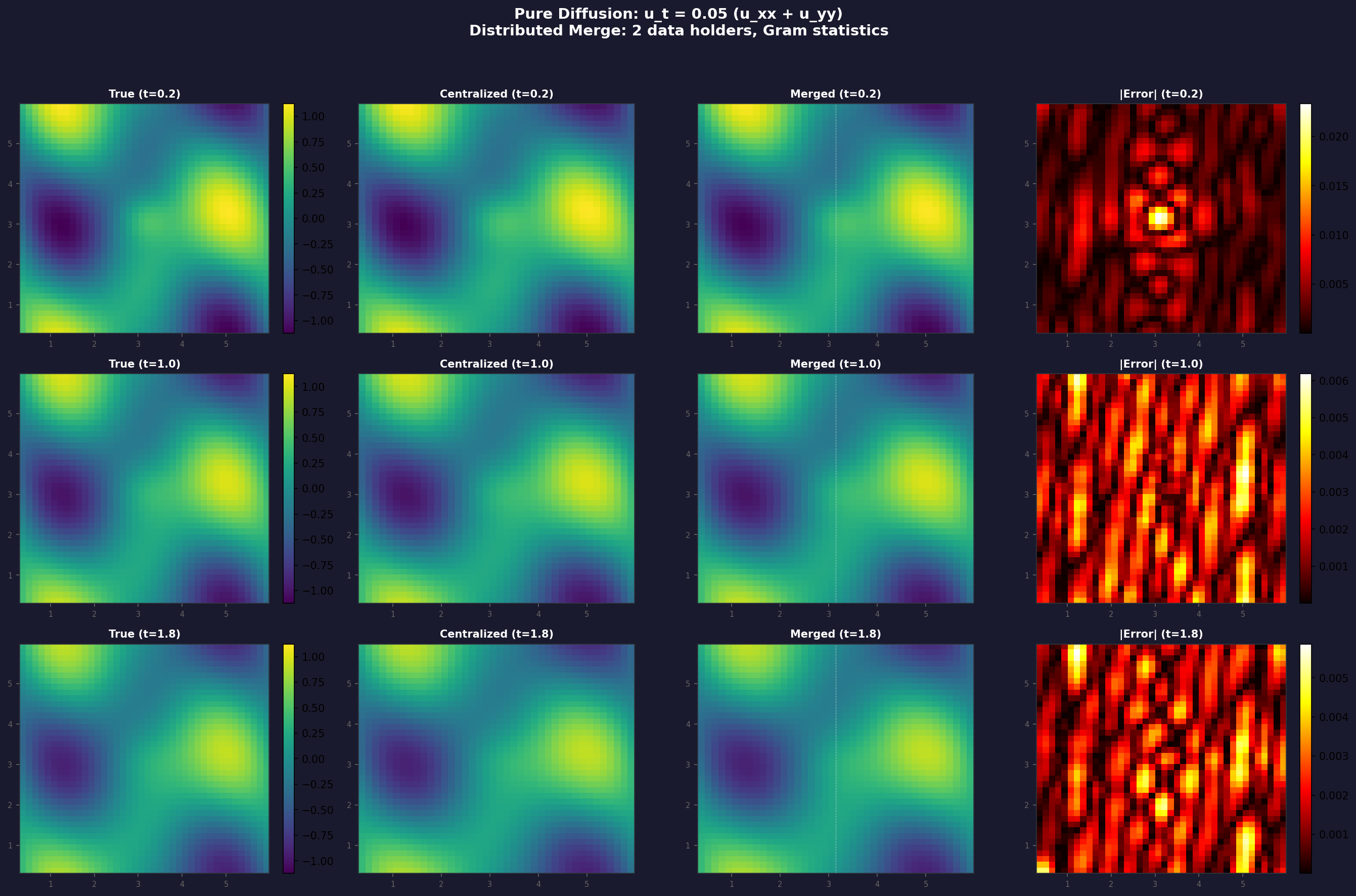}
\caption{Diffusion: $u_t = D\nabla^2 u$ ($D{=}0.05$). The field smooths as heat diffuses.}
\label{fig:diffusion}
\end{subfigure}
\vspace{0.3em}
\begin{subfigure}[b]{\linewidth}
\includegraphics[width=\linewidth]{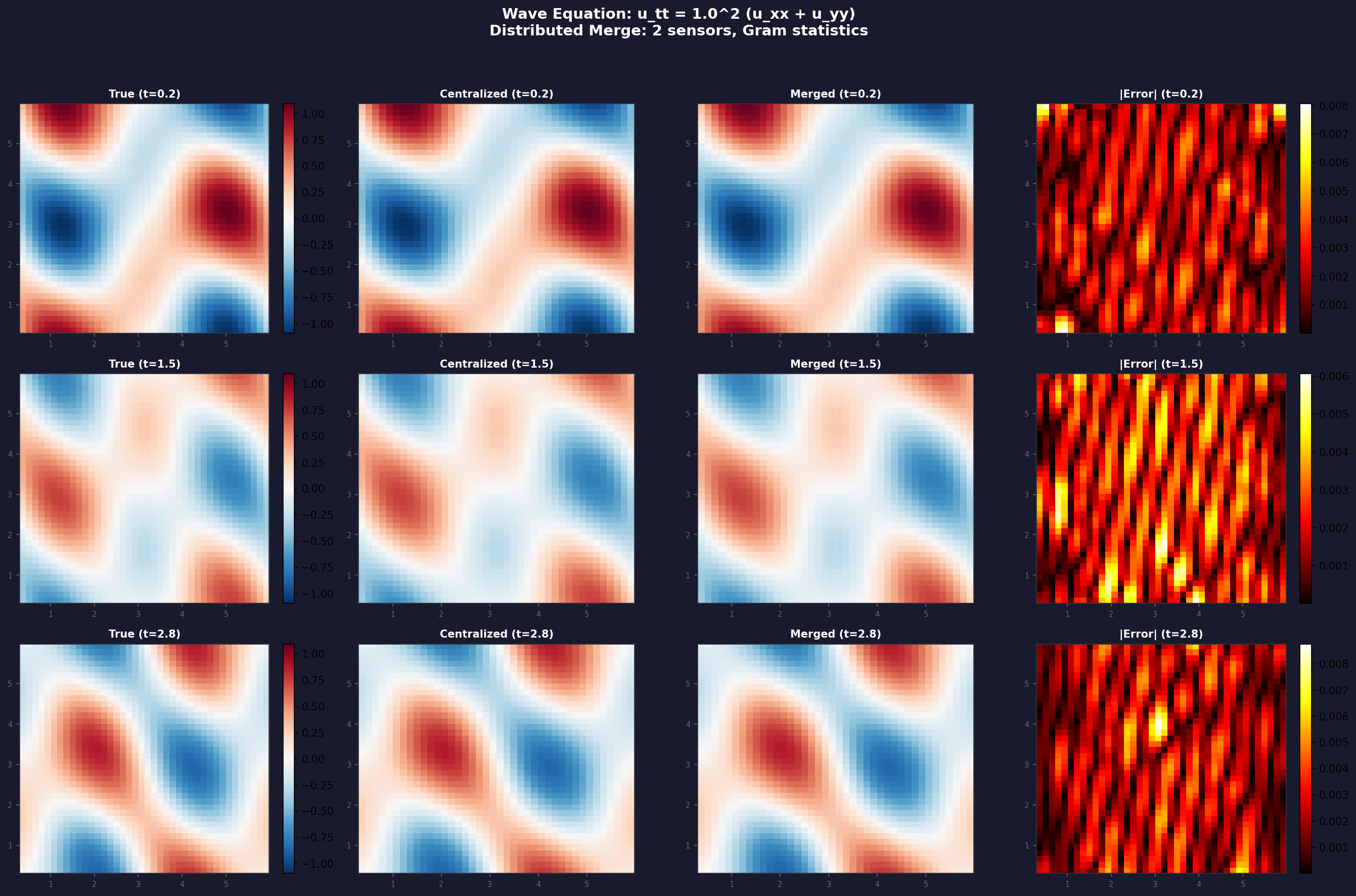}
\caption{Wave: $u_{tt} = c_w^2 \nabla^2 u$ ($c_w{=}1$). Energy is conserved; the field persists.}
\label{fig:wave}
\end{subfigure}
\end{center}
\caption{Synthetic 2D PDE experiments. Each row shows ground truth, centralized fit, Gram merge (two data holders, left/right split at dashed line), and absolute error at three time steps. Centralized and merged reconstructions are identical ($\delta_{\max} < 10^{-12}$). The Burgers experiment (1D, nonlinear) is in Appendix~\ref{app:figures}.}
\label{fig:pde}
\end{figure}

Figure~\ref{fig:pde} shows both 2D PDE experiments. The diffusion field (Figure~\ref{fig:diffusion}) smooths progressively; the wave field (Figure~\ref{fig:wave}) conserves energy. In both cases, centralized and merged reconstructions are visually indistinguishable; the error panels reflect field-approximation error, not the merge.

\subsection{Wave Equation}

The wave equation $u_{tt} = c_w^2(u_{xx} + u_{yy})$ with $c_w = 1$ is solved by analytic evolution of its Fourier modes. From 10,000 observations ($K_x{=}11, K_y{=}11, K_t{=}13$; $P{=}1{,}573$), the recovered wave speed is $c_{w,\mathrm{rec}} = 0.9991$ (0.09\% error). The distributed merge matches centralized to $\delta_{\max} < 10^{-12}$.

\subsection{Viscous Burgers Equation (Nonlinear)}

The 1D viscous Burgers equation $u_t + u\,u_x = \kappa\, u_{xx}$ with $\kappa = 0.1$ is solved via pseudospectral methods. The domain is $[0, 2\pi] \times [0, 1]$, fitted with a 2D basis ($K_x{=}15, K_t{=}13$; $P{=}195$). With 5,000 observations, $\kappa_{\mathrm{rec}} = 0.1050$ (5.0\% error). The Burgers experiment introduces the nonlinear feature $u\,u_x$ and has higher parameter error in this configuration; the contributions of representation and derivative-estimation errors have not been isolated. A grid-convergence check on the reference field shows that refining from $80{\times}60$ to $320{\times}240$ reduces reference-field recovery error from 0.89\% to 0.06\%, indicating finite-difference discretization error in the reference evaluation. The merge reproduces centralized exactly ($\delta_{\max} < 2 \times 10^{-11}$).

\subsection{Heat Equation with Spatially Varying Source}

The inhomogeneous heat equation $u_t = D(u_{xx} + u_{yy}) + S(x,y)$ with $D = 0.03$ and a periodic source $S(x,y) = 0.3 + 0.2\cos x + 0.1\sin 2y$ is solved exactly via the modal formula $\hat{u}_k(t) = e^{-a_k t}\hat{u}_k(0) + (1 - e^{-a_k t})\hat{S}_k / a_k$ for $|k|>0$ (where $a_k = D|k|^2$) and $\hat{u}_0(t) = \hat{u}_0(0) + t\hat{S}_0$ for the zero mode. With 10,000 observations, $D_{\mathrm{rec}} = 0.02977$ (0.76\% relative error from unrounded value). The distributed merge matches centralized to $\delta_{\max} < 10^{-12}$.

\subsection{Noise Sensitivity and Robustness}

A systematic noise sensitivity study (Appendix Table~\ref{tab:noise}) crosses three observation densities (1K, 3K, 10K), three noise levels (0\%, 1\%, 5\%), and five independent seeds for both diffusion and wave. Mean relative parameter error across five seeds is below 1\% at 3,000+ observations with up to 1\% noise. The distributed merge reproduces the centralized estimator in all 90 conditions ($\delta_{\max} < 10^{-11}$). Signed-bias analysis reveals systematic attenuation of diffusion estimates under high noise, consistent with errors-in-variables effects. Note that with a fixed unnormalized ridge penalty $\lambda$, increasing $N$ changes both data density and effective regularization strength; this confound has not been controlled for.

\begin{table}[t]
\caption{Distributed merge: Gram merge versus centralized fitting on balanced regional draws (5,000 per holder for 2D PDEs, 2,500 for Burgers). $\delta_{\max}$: maximum absolute prediction difference on the evaluation grid. Main-text parameter estimates use separate full-domain draws, explaining slight differences.}
\label{tab:pde}
\begin{center}
\small
\begin{tabular}{lcccc}
\toprule
PDE & Method & RMSE & Recovered Parameter & $\delta_{\max}$ \\
\midrule
\multirow{2}{*}{Diffusion}
 & Centralized & 0.00252 & $D_{\mathrm{rec}} = 0.04989$ & \\
 & Gram merge & 0.00252 & $D_{\mathrm{rec}} = 0.04989$ & $< 10^{-12}$ \\
\midrule
\multirow{2}{*}{Wave}
 & Centralized & 0.00178 & $c_{w,\mathrm{rec}} = 0.9991$ & \\
 & Gram merge & 0.00178 & $c_{w,\mathrm{rec}} = 0.9991$ & $< 10^{-12}$ \\
\midrule
\multirow{2}{*}{Burgers}
 & Centralized & 0.01096 & $\kappa_{\mathrm{rec}} = 0.1050$ & \\
 & Gram merge & 0.01096 & $\kappa_{\mathrm{rec}} = 0.1050$ & $< 2 \times 10^{-11}$ \\
\midrule
\multirow{2}{*}{Heat+source}
 & Centralized & 0.00095 & $D_{\mathrm{rec}} = 0.02995$ & \\
 & Gram merge & 0.00095 & $D_{\mathrm{rec}} = 0.02995$ & $< 10^{-12}$ \\
\bottomrule
\end{tabular}
\end{center}
\end{table}

Table~\ref{tab:pde} summarizes the distributed merge results across all four PDEs. The dense reconstruction experiments use a full-domain random draw; the distributed experiments use a balanced regional draw with a different sampling procedure, which is why parameter estimates differ slightly between the main text and the table. Individual data holders fail as extrapolators outside their observed regions, but the merged field recovers the full domain. An independent verification applying the same finite-difference stencils to exact analytic field values (without spline fitting) yields 0.13\% wave-speed error and 0.45\% diffusion error from the derivative grid alone. These controls quantify a nonzero discretization contribution; they are not lower bounds on full-pipeline parameter error, because reconstruction and differentiation errors may reinforce or partially cancel.

\subsection{Real Data: NOAA Sea-Surface Temperature}
\label{sec:noaa}

To test the framework beyond synthetic settings, it is applied to National Oceanic and Atmospheric Administration (NOAA) OI SST V2 monthly sea-surface temperature data \citep{reynolds2002noaa}, covering December 1981 to January 2023 over the tropical Pacific ($30^\circ$S--$30^\circ$N, $120^\circ$E--$280^\circ$E). The representation uses 2D spatial B-splines ($K_\text{lat} = 13$, $K_\text{lon} = 23$) combined with temporal Fourier harmonics and polynomial trend terms, totaling 2,093 parameters fitted to approximately 4.41 million ocean grid-month values (after excluding land cells using the product's land--sea mask).

\begin{figure}[t]
\begin{center}
\includegraphics[width=\linewidth]{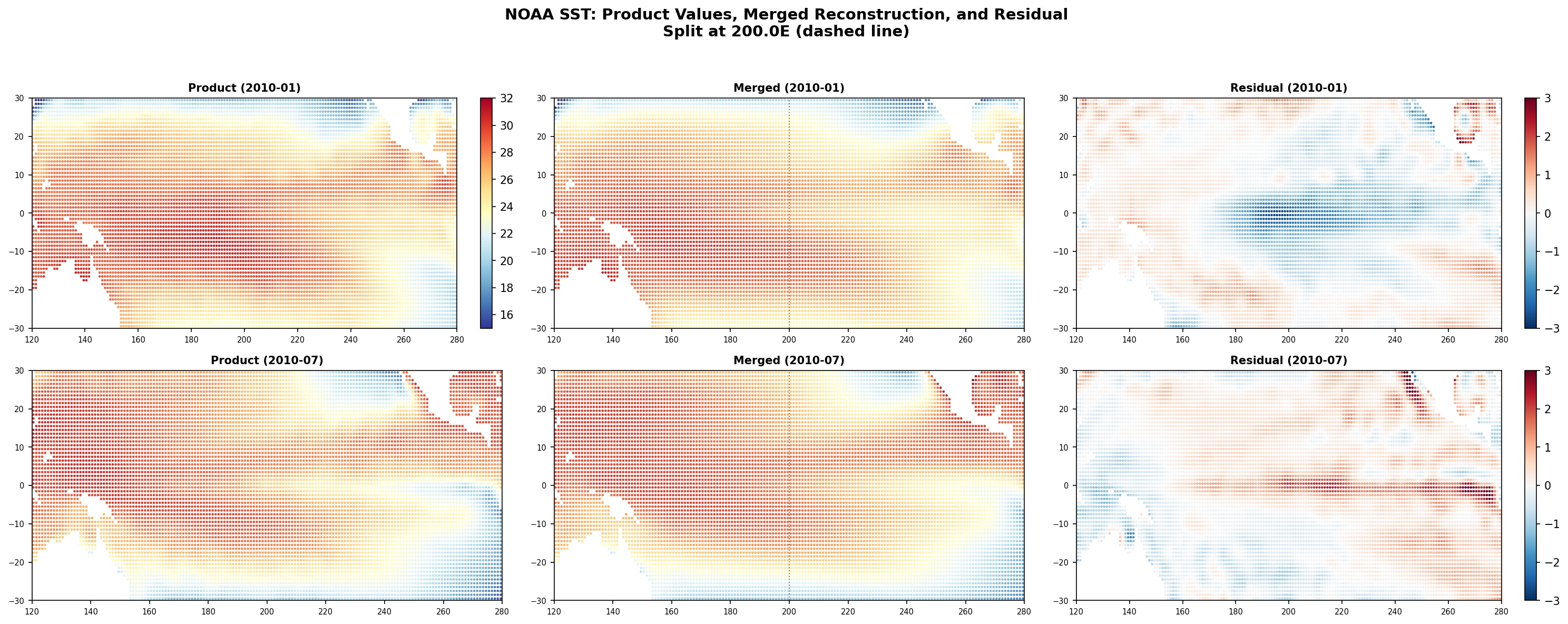}
\end{center}
\caption{NOAA SST: product values (left), merged reconstruction from two data holders split at 200$^\circ$E (center, dashed line), and reconstruction residual (reconstruction minus product, in $^\circ$C, right) for January and July 2010. This is reconstruction error, not merged-versus-centralized disagreement.}
\label{fig:sst}
\end{figure}

Figure~\ref{fig:sst} shows the merged reconstruction for two months. Partitioning into west (120--200$^\circ$E) and east (200--280$^\circ$E) data holders, the Gram merge matches centralized fitting (in-sample RMSE 0.639$^\circ$C both). Held-out evaluation withholds 20\% of ocean points (spatial, seed 42) or every 6th month (temporal); held-out RMSE is 0.643$^\circ$C and 0.663$^\circ$C respectively (using $\lambda{=}10^{-4}$; the full-data experiment uses $\lambda{=}10^{-1}$), supporting spatial and temporal interpolation on this product. The merge reproduces centralized reconstruction RMSEs in both held-out tests. Note that this partitions a gridded analysis product \citep{noaa_psl_oisstv2}, not raw sensor records.

\section{Conclusion}

Tensor-product spline fields admit composable sufficient statistics, enabling exact distributed merging and physical inference from fragmented observations. Validated on four PDEs and real climate data, the merge introduces zero degradation: prediction-space discrepancies remain below $2 \times 10^{-11}$ across all synthetic experiments, with sub-percent parameter recovery for linear PDEs and 5\% for nonlinear Burgers. For NOAA SST, merged and centralized reconstruction RMSEs match to displayed precision. For fixed-feature squared-error fitting, the protocol requires communicating only Gram matrices and moment vectors. Limitations include tensor-product scaling ($P = k^q$), finite-difference derivatives, and restriction to known PDE structure. See Appendix~\ref{app:discussion} for further discussion.

\subsection*{Reproducibility Statement}

All experiments use fixed random seeds. Diffusion, wave, and heat-with-source solutions are generated by analytic evolution of represented Fourier modes. Burgers solutions use pseudospectral integration (dealiased 3/2 rule, RK4, $N{=}256$, $\Delta t{=}0.001$). Observation evaluation and derivative estimation introduce additional numerical approximations, assessed through reference tests described in Section~4. The NOAA SST data is publicly available from NOAA/OAR/ESRL PSL. The InKAN B-spline package is publicly available \citep{mysore2026inkan}. Experiment code is available at the repository linked in the abstract. Key hyperparameters (grid sizes, regularization strength, observation counts) are reported in each experiment description. Complete specifications, including initial conditions, coordinate normalization, spline knot construction, NOAA temporal basis definitions, noise distributions, and random seeds, are provided in the supplementary code.

\subsection*{Ethics Statement}

This work involves no human subjects or private data. The NOAA SST dataset is publicly available. The distributed merging framework communicates sufficient statistics (Gram matrices and moment vectors) rather than individual observation records. Privacy leakage through the shared statistics is not analyzed in this work and no formal privacy guarantee is claimed.

\subsection*{AI Use Statement}

The research concept, experimental design, and scientific analysis are the author's own work. The author wrote the core implementation, ran all experiments, and performed the scholarly investigation including primary-source verification of all citations. Claude (Anthropic) and ChatGPT (OpenAI) were used as assistive tools for code debugging, literature search, prose editing, and mathematical cross-checks. The author takes responsibility for the final methods, results, and text.

\bibliography{references}
\bibliographystyle{iclr2027_conference}

\raggedbottom
\newpage
\appendix
\section{Experimental Configuration and Additional Results}
\label{app:figures}

All experiments use InKAN v0.4.2 (piecewise polynomial basis), PyTorch (float64), and were developed on NVIDIA H100 80GB GPUs; the released code runs on CPU. Table~\ref{tab:config} summarizes the configuration.

\vspace{0.5em}
\noindent
\begin{minipage}{\linewidth}
\captionof{table}{Experimental configuration. $P$: total coefficients. $N_{\text{eval}}$: FD derivative grid.}
\label{tab:config}
\centering
\scriptsize
\begin{tabular}{lccccc}
\toprule
& Diffusion & Wave & Burgers & Heat+source & NOAA SST \\
\midrule
PDE & $u_t{=}D\nabla^2 u$ & $u_{tt}{=}c_w^2\nabla^2 u$ & $u_t{+}uu_x{=}\kappa u_{xx}$ & $u_t{=}D\nabla^2 u{+}S$ & (fitting) \\
Domain & $[0,2\pi]^2{\times}[0,2]$ & $[0,2\pi]^2{\times}[0,3]$ & $[0,2\pi]{\times}[0,1]$ & $[0,2\pi]^2{\times}[0,2]$ & Pacific, 41yr \\
True param. & $D{=}0.05$ & $c_w{=}1$ & $\kappa{=}0.1$ & $D{=}0.03$ & -- \\
Basis & $11^3$ & $11^2{\times}13$ & $15{\times}13$ & $11^3$ & $13{\times}23{\times}7$ \\
$P$ & 1331 & 1573 & 195 & 1331 & 2093 \\
$N_{\text{eval}}$ & $40^2{\times}30$ & $40^2{\times}40$ & $80{\times}60$ & $40^2{\times}30$ & -- \\
$N_{\text{obs}}$ & 10K & 10K & 5K & 10K & 4.41M \\
\bottomrule
\end{tabular}
\end{minipage}

\vspace{0.5em}
\noindent
\begin{minipage}{\linewidth}
\centering
\includegraphics[width=0.95\linewidth]{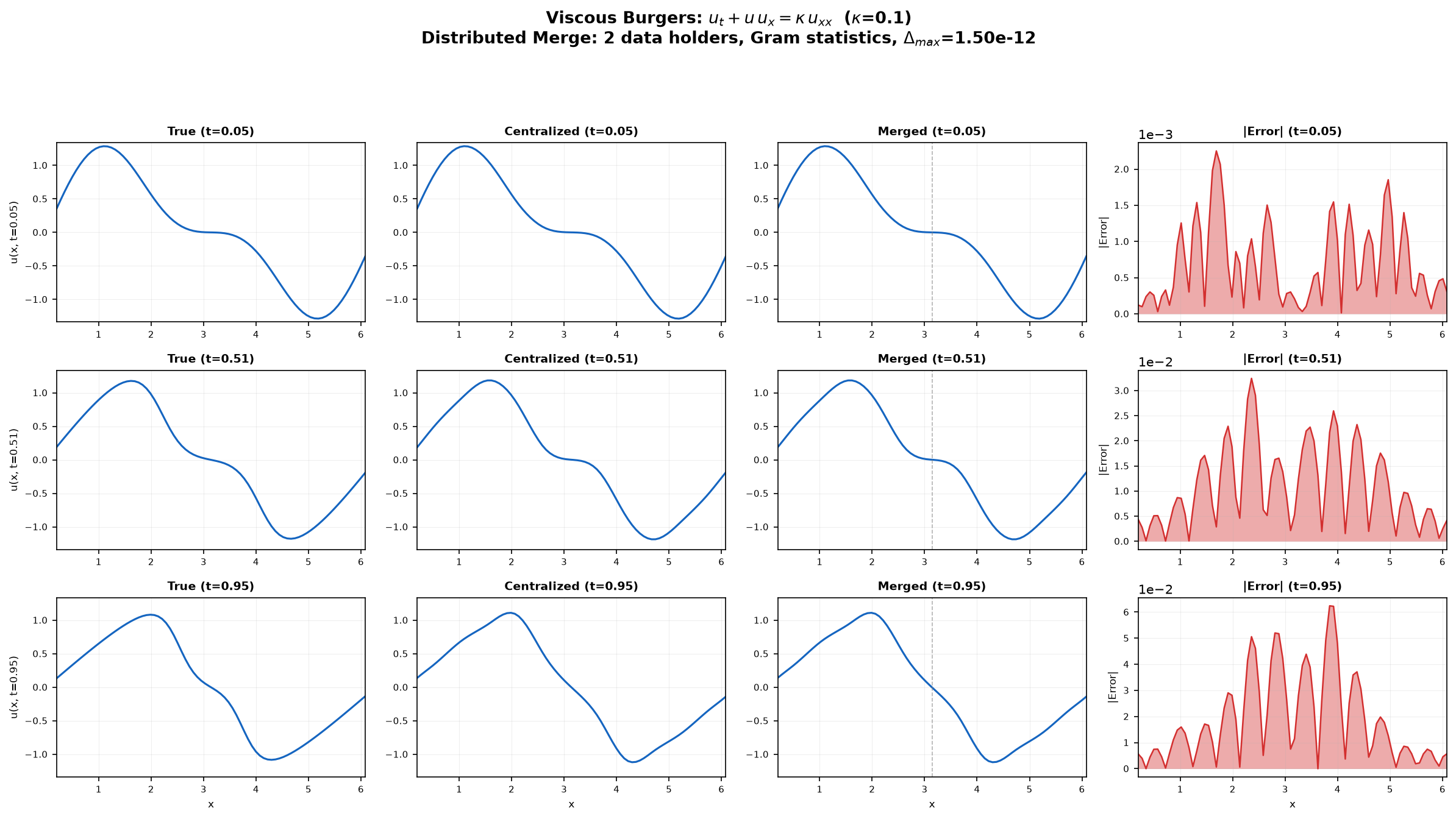}
\captionof{figure}{Viscous Burgers $u_t + u\,u_x = \kappa\, u_{xx}$: ground truth, centralized fit, Gram merge, and absolute error. The merged field matches centralized ($\delta_{\max} < 2 \times 10^{-11}$).}
\label{fig:burgers}
\end{minipage}

\vspace{0.5em}
\noindent
\begin{minipage}{\linewidth}
\centering
\includegraphics[width=0.95\linewidth]{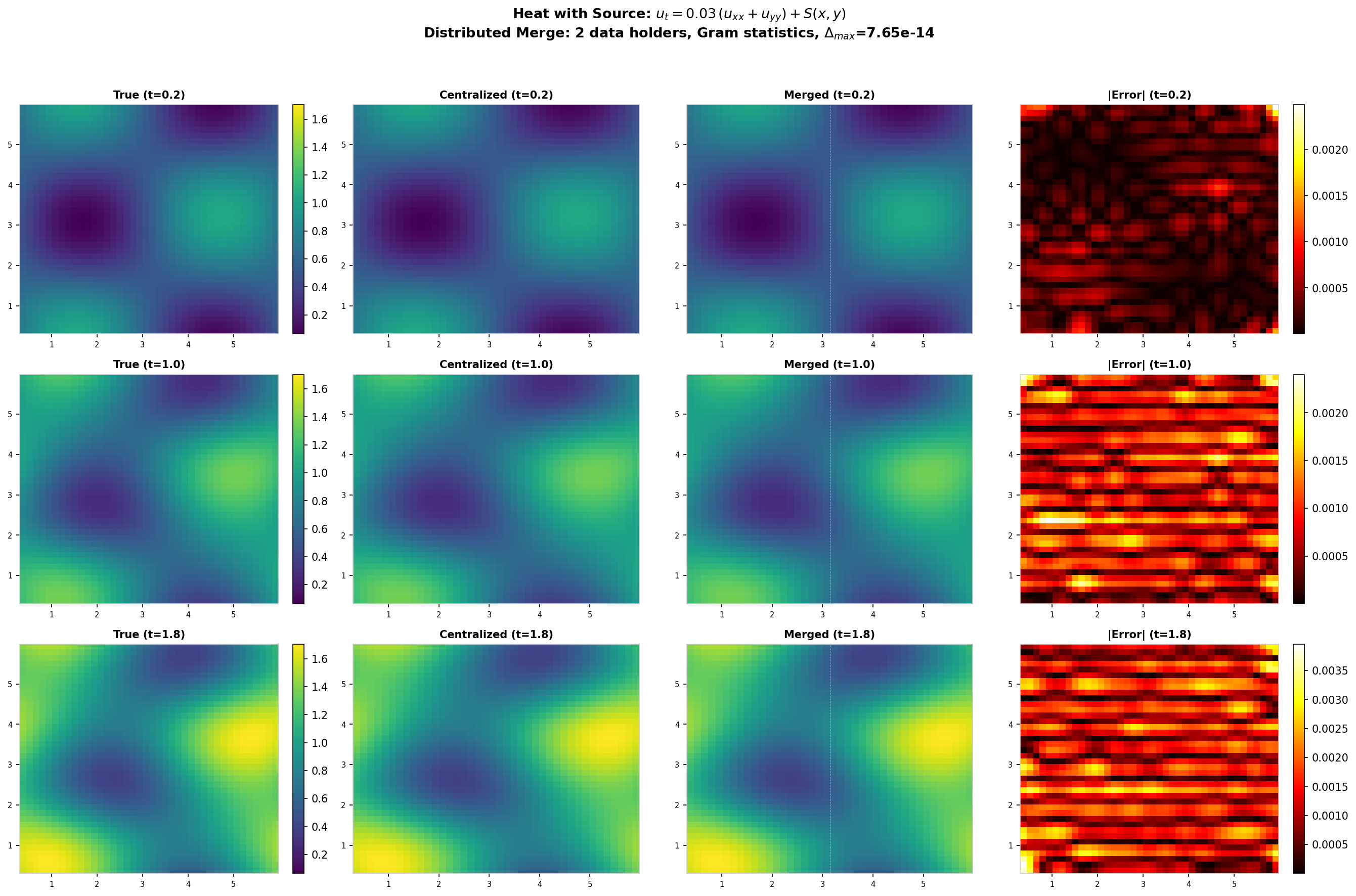}
\captionof{figure}{Heat with source $u_t = D\nabla^2 u + S(x,y)$: ground truth, centralized fit, Gram merge, and absolute error. The merged field matches centralized ($\delta_{\max} < 10^{-12}$).}
\label{fig:heat_source}
\end{minipage}

\section{Noise Sensitivity Study}
\label{app:noise}

\begin{table}[H]
\caption{Parameter recovery: mean $\pm$ std of relative error (\%) over 5 independent seeds. Noise levels are defined relative to a fixed reference-grid field standard deviation. Sample standard deviations use $\text{ddof}{=}1$. The distributed merge reproduces the centralized estimator in all 90 conditions ($\delta_{\max} < 10^{-11}$). All wave conditions produced valid $\widehat{c_w^2} \geq 0$.}
\label{tab:noise}
\begin{center}
\scriptsize
\begin{tabular}{lcccccc}
\toprule
& \multicolumn{3}{c}{Diffusion ($D{=}0.05$)} & \multicolumn{3}{c}{Wave ($c_w{=}1$)} \\
\cmidrule(lr){2-4} \cmidrule(lr){5-7}
$N_{\text{obs}}$ & 0\% noise & 1\% noise & 5\% noise & 0\% noise & 1\% noise & 5\% noise \\
\midrule
1,000 & $2.3 \pm 2.3$ & $2.0 \pm 2.6$ & $8.1 \pm 5.3$ & $0.9 \pm 0.6$ & $0.6 \pm 0.5$ & $8.4 \pm 0.8$ \\
3,000 & $0.3 \pm 0.2$ & $0.4 \pm 0.2$ & $2.6 \pm 1.7$ & $0.4 \pm 0.04$ & $0.3 \pm 0.2$ & $1.2 \pm 0.9$ \\
10,000 & $0.1 \pm 0.08$ & $0.4 \pm 0.4$ & $1.8 \pm 1.9$ & $0.04 \pm 0.03$ & $0.06 \pm 0.03$ & $0.5 \pm 0.4$ \\
\bottomrule
\end{tabular}
\end{center}
\end{table}

Signed-bias analysis shows diffusion estimates are systematically attenuated under high noise (mean bias $-0.003$ at 5\% noise, 1K observations), consistent with errors-in-variables effects. Wave-speed recovery is more robust at high observation counts because the estimator depends on $u_{tt}$ and the Laplacian, both of which benefit from temporal smoothing in the spline fit.

\section{Discussion}
\label{app:discussion}

\textbf{Scope of exactness.} The exact merge holds for fixed-feature models optimized under the quadratic objective in Eq.~(\ref{eq:loss}). It does not hold for multi-layer networks where independently trained hidden representations diverge, because changing an earlier layer invalidates the Gram matrices computed at later layers. The composability guarantee extends to any globally shared quadratic penalty $\lambda\, c^\top R\, c$ (e.g., derivative-based smoothing penalties), provided $\sum_s G_s + \lambda R$ is positive definite.

\textbf{Basis geometry.} Basis support constrains Gram sparsity, but averaging performance additionally depends on function space, coverage, regularization, and conditioning. A controlled basis-family comparison is left to future work.

\textbf{Scaling and limitations.} The coefficient count $P = k^q$ grows exponentially with input dimension; dense Gram storage is $O(P^2)$ and the solve is $O(P^3)$. In the synthetic experiments, dense summaries exceed raw-data size; communication savings require sparsity exploitation not implemented here. Finite-difference derivative estimation limits physics recovery accuracy. The NOAA SST experiment uses an already-interpolated product; real-data physical-parameter inference remains outside the validated scope. Privacy leakage through the shared statistics is not analyzed.

\textbf{Future directions.} Analytic B-spline derivatives, periodic basis functions, vector-field extensions, and quantitative comparison with federated scientific machine learning methods \citep{zhang2025fedsciml} are natural next steps.

\end{document}